\documentclass[10pt,twocolumn,letterpaper]{article}

\usepackage[pagenumbers]{wacv} 

\usepackage[accsupp]{axessibility}  

\usepackage[pagebackref,breaklinks,colorlinks,allcolors=iccvblue]{hyperref}
\usepackage[textsize=tiny]{todonotes}
\usepackage[detect-all,locale=UK,per-mode=symbol]{siunitx}
\DeclareSIUnit\pixel{px}
\usepackage[accsupp]{axessibility}  
\usepackage{microtype}
\usepackage{algorithm}
\usepackage{algorithmic}
\usepackage{colortbl}
\usepackage{multirow}
\usepackage{tabularx}
\usepackage{adjustbox}
\usepackage{graphicx}
\usepackage{amsmath}
\usepackage{amssymb}
\usepackage{booktabs}
\usepackage{tikz}
\usepackage{xspace}
\newcommand\ours{\text{FlowEO}\xspace}
\definecolor{customgreen}{RGB}{193,199,205}
\definecolor{customblue}{RGB}{166,200,254}
\definecolor{iccvblue}{rgb}{0.21,0.49,0.74}
\usepackage[pagebackref,breaklinks,colorlinks]{hyperref}

\usepackage[capitalize]{cleveref}
\crefname{section}{Sec.}{Secs.}
\Crefname{section}{Section}{Sections}
\Crefname{table}{Table}{Tables}
\crefname{table}{Tab.}{Tabs.}

\def\wacvPaperID{870} 
\def\confName{WACV}
\def\confYear{2026}

\begin{document}

\title{FlowEO: Generative Unsupervised Domain Adaptation for Earth Observation}

\author{Georges Le Bellier \textsuperscript{1}\\
\textsuperscript{1} Cnam, CEDRIC, EA4629\\
F-75141 Paris, France\\
{\tt\small georges.le-bellier@lecnam.net}
\and
Nicolas Audebert \textsuperscript{1,2}\\
\textsuperscript{2} Univ. Gustave Eiffel, ENSG, IGN, LASTIG\\ F-94160 Saint-Mandé, France\\
{\tt\small nicolas.audebert@ign.fr}
}

\maketitle


\begin{abstract}

The increasing availability of Earth observation data offers unprecedented opportunities for large-scale environmental monitoring and analysis. However, these datasets are inherently heterogeneous, stemming from diverse sensors, geographical regions, acquisition times, and atmospheric conditions. 
Distribution shifts between training and deployment domains severely limit the generalization of pretrained remote sensing models, making unsupervised domain adaptation (UDA) crucial for real-world applications.
We introduce \ours{}, a novel framework that leverages generative models for image-space UDA in Earth observation. We leverage flow matching to learn a semantically preserving mapping that transports from the source to the target image distribution. This allows us to tackle challenging domain adaptation configurations for classification and semantic segmentation of Earth observation images. We conduct extensive experiments across four datasets covering adaptation scenarios such as SAR to optical translation and temporal and semantic shifts caused by natural disasters. 
Experimental results demonstrate that \ours{} outperforms existing image translation approaches for domain adaptation while achieving on-par or better perceptual image quality, highlighting the potential of flow-matching-based UDA for remote sensing.

\end{abstract}

\section{Introduction}

\begin{figure}[h!]
    \hspace*{-15px} 
    \centering
    \includegraphics[width=0.92\linewidth, trim=0px 0px 0px 5px, clip]{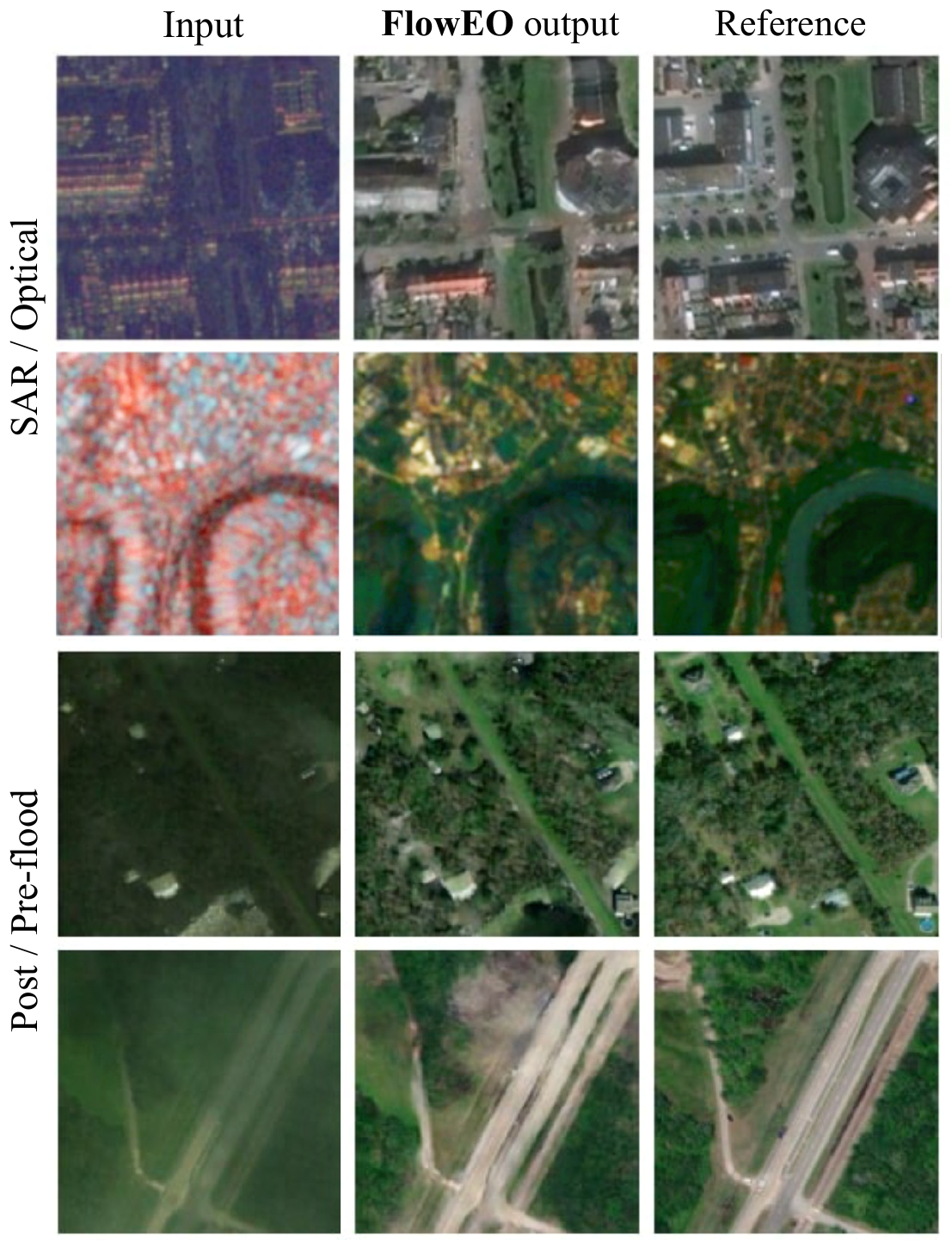} 
    \caption{\textbf{\ours{}} generates realistic and semantically consistent outputs on various challenging image translation tasks, such as pre- to post-disaster domain adaptation and SAR-to-Optical translation.}
    \label{fig:framework_overview}
    \vspace{-6mm}
\end{figure}

Large amounts of remote sensing images are collected at high frequency to analyze and model the complexity of physical phenomena on Earth. The diversity of the data acquired calls into question the use of pretrained models to process them. Indeed, the phenomena studied on the Earth's surface are non-stationary and subject to great variability due to seasonal variations, human-made changes, and extreme events such as wildfires and floods. This causes drifts in the data distribution, compromising model performance at inference~\cite{kelleberger2021deep}. In addition, sensors with complementary characteristics are used to capture multiple views of the same area and overcome sensor limitations, \eg ground occultation by clouds for optical sensors can be alleviated using radar.

Emergency management of natural disasters requires rapid analysis of the ground-level situation to plan rescue operations and assess environmental consequences. However, the domain shift between post-disaster and ordinary satellite images degrades the performance of off-the-shelf deep models. The urgency of such events makes it impossible to annotate a dataset for supervised training in disaster-affected areas. Domain adaptation is therefore a promising solution to speed up image analysis for disaster management.

Similarly, robust remote sensing pipelines leverage the strengths of all available Earth observation sensors. For example, Synthetic Aperture Radar (SAR) provides all-weather day-and-night imaging capabilities, as its wavelength allows it to penetrate clouds and operate independently of illumination conditions. However, due to their speckle noise and sensitivity to terrain geometry, interpreting SAR images is harder for humans than optical images \cite{tupin:hal-02287185}. Therefore, cross-sensor domain adaptation has been well-investigated in Earth observation. SAR-to-Optical translation (S2O) in particular can provide human-interpretable optical images in contexts where only SAR imagery is available \cite{s2opotential,sartooptical_nature, 8825802}, \eg to fill in missing optical due to cloud cover. This provides higher frequency images by leveraging co-located multi-modal SAR/optical acquisitions. In turn, this enables disaster monitoring and environmental surveillance in scenarios where acquiring cloud-free optical data is challenging. 

Major efforts have been made in recent years to overcome distribution drift through domain adaptation \cite{pizzati2020domain, kelleberger2021deep, 7486184}. Due to the low availability of labeled satellite image datasets, \emph{unsupervised} domain adaptation methods have been preferred as they only require labels in the source domain. 
Unsupervised domain adaptation is mainly studied inside the feature space of a pretrained model \cite{Xia_2021_ICCV, damodaran2018deepjdot}. Because of the lower dimensionality of the latent space, this favours classification tasks \cite{pmlrle21am, Gilo2023RDAOTRU}, although some approaches also have been proposed for dense tasks such as segmentation \cite{9157228, Choe_2024_CVPR, 10120939}. To overcome this limitation, domain adaptation can be applied in image-space \cite{zhao2019multi}. It facilitates the transfer interpretation and improves explainability while disentangling transfer and downstream tasks. Such image translation approaches are orthogonal to future improvements in classifiers and can be used with any inference model without retraining. To this end, we employ flow matching models \cite{lipman2022flow, albergo2023stochastic, peluchetti2023non}, a new family of models that have demonstrated high-quality generation across various modalities \cite{esser2024scaling, wohlwend2024boltz}.

\noindent\textbf{\ours{}.} We propose \ours{}, a new model that leverages flow matching models for unsupervised domain adaptation in Earth Observation. We introduce a novel domain adaptation method in pixel-space, enabling visual interpretation, and test it extensively on four datasets covering classification and segmentation tasks, demonstrating its effectiveness for dense downstream tasks in challenging scenarios of post-disaster domain adaptation and sensor translation. In summary:
\begin{enumerate}
    \item We introduce \ours{}, a new generative UDA method, downstream-task-agnostic that does not require modification or retraining of downstream predictive models. 
    \item We are the first to leverage latent flow matching for data-to-data translation, on multiple remote sensing modalities, including SAR, low-resolution, and high-resolution optical data.
    \item We introduce an application-driven evaluation protocol, going beyond standard image generation metrics to assess the impact of UDA on real-world Earth observation tasks: semantic segmentation and classification.
\end{enumerate}

\section{Related Work}

\begin{figure*}[t]
\vskip 0.2in
\begin{center}
\centerline{\includegraphics[width=1\textwidth, trim=15px 0px 15px 0px, clip]{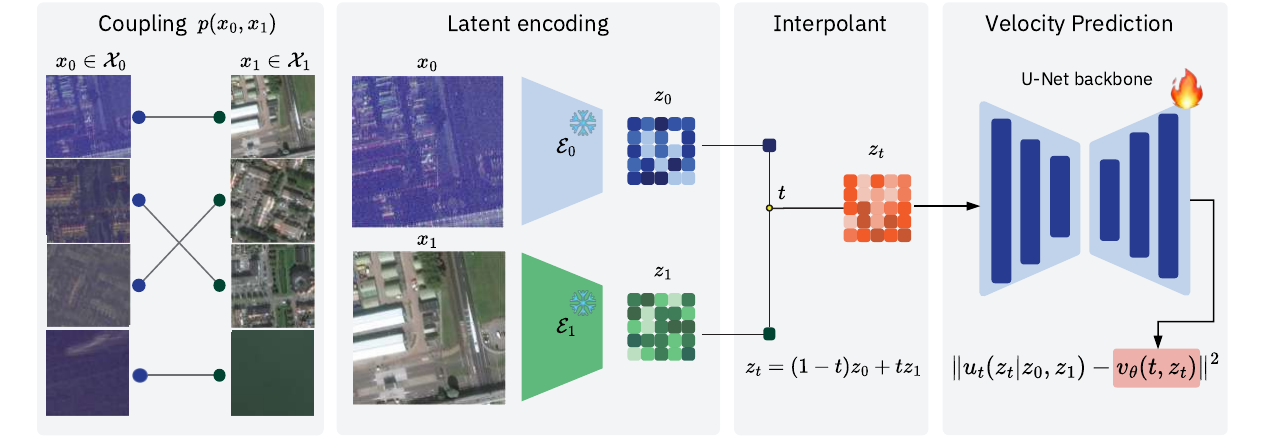}}
\caption{\ours{} learns a latent flow between the source and target distributions in four stages: 1) the training image pairs are sampled from the coupling $p(x_0, x_1)$, 2) images are encoded in SD3 latent space, 3) we interpolate between the latent codes $z_0$ and $z_1$ to compute $z_t$ for $t\sim \mathcal{U}(0, 1)$, 4) we train the U-Net backbone $v_\theta$ on a simple regression loss to match the conditional velocity $u_t(z_t\mid z_0, z_1)$.}
\label{fig:main_figure}
\end{center}
\vspace{-10mm}
\end{figure*}

\subsection{Unsupervised domain adaptation}

Consider two distinct domains, represented by two datasets $D_0$ and $D_1$. Suppose that one contains annotations, \ie $D_1 = \{\mathcal{X}_1, \mathcal{Y}_1\}$. This is the \emph{source} domain, on which a predictive model $S_1$ has been trained, \eg for segmentation, classification, regression, etc. Conversely, let $D_0 = \{\mathcal{X}_0, \emptyset\}$ be the unlabeled \emph{target} domain, on which we would like to infer new predictions. The absence of annotations on $D_0$  prevents us from training a predictive model on it. Instead, we intend to use the existing model $S_1$ for the new $D_0$ data. However, the underlying differences between the two domains will result in a drop in its performance if applied directly to the new domain. Its generalization capabilities do not allow direct transfer of segmentation scores. Domain adaptation aims to extend a model's performance beyond its training domain by means of an adaptation procedure. 

Domain adaptation techniques are split into two broad families. First, domain adaptation can be applied post-hoc on an existing predictive model. These approaches aim to align the features obtained from the predictive model, \eg with optimal transport \cite{damodaran2018deepjdot,Gilo2023RDAOTRU}, or fine-tuning/adapting the weights of the model to the new domain \cite{905255,Xia_2021_ICCV}. However, every downstream model needs to be adapted, which can be costly and constrains usage of “off-the-shelf” models. Second, adaptation can take place directly in the data space, \ie image space in our case. Instead of adapting the model to the target domain, the target data is altered to match the source domain. This approach, called image-to-image translation for domain adaptation \cite{murez_image_2018}, leverages conditional generative models derived from style transfer \cite{jing2019neural}.

\subsection{Image translation for domain adaptation}
Image translation builds upon the seminal work of Pix2Pix \cite{isola2017image}, that trains an image-to-image model on paired datasets using a combination of supervised regression loss and an adversarial loss using a patch-wise GAN. It has been extended to the unpaired setting into CycleGAN \cite{CycleGAN2017}, leveraging cycle consistency by training two GANs in symmetry. These models have been used for domain adaptation in multiple settings, including dehazing \cite{shao_domain_2020}, tactile perception \cite{jing_unsupervised_2023}, and semantic segmentation \cite{xie_selfsupervised_2020}. More recent models include StegoGAN \cite{wu2024stegogan} that explicitly deals with features that are impossible to match between the two domains, and more recent generative model classes, \eg diffusion models and Schrödinger bridges \cite{de2021diffusion, shi2024diffusion, zhou2023denoising}. The Unpaired Neural Schrödinger Bridge \cite{kim2023unsb}, for example, has found success for domain adaptation of medical CT scans \cite{teimouri_ctbased_2024}.

\medskip
\noindent\textbf{Image translation for Earth Observation}
Such approaches are also common in Earth Observation. For example, StegoGAN \cite{wu2024stegogan} performs style transfer from satellite images to maps and vice-versa. Natural disaster management is also the subject of domain adaptation research, via flood simulation using adversarial networks conditioned on physical measurements \cite{10758300}. SAR to optical (S2O) image translation is especially popular because radar acquisitions can be carried out despite cloud cover, optical sensors suffer from cloud occultation. S2O imagery makes it possible to fill missing optical acquisitions based on SAR images from close dates. Research has leveraged various classes of generative models for S2O, \eg \cite{rs15071878} uses a conditional GAN, \cite{segcyclegan} uses CycleGAN, \cite{kim2024conditional} uses diffusion bridges, and so on.
\ours pushes forward this state-of-the-art by integrating flow matching models that deliver a better semantic-preserving transfer and higher quality generation. 

\medskip
\textbf{Flow Matching}
\emph{Flow matching models} (FMMs) have been introduced in the last years \cite{lipman2022flow, albergo2023stochastic, peluchetti2023non} and now represent the state of the art in generative models for various applications \cite{esser2024scaling, ma2024sit, wohlwend2024boltz}. 
However, flow matching models also allow data-to-data transport between arbitrary distributions \cite{albergo2023couplings, liu2022rectified}, and remain less well-studied. Contrary to diffusion bridges \cite{zhou2023denoising, albergo2023stochastic} and Schrödinger Bridges \cite{bortoli2024schrodinger, de2021diffusion} that rely on stochastic differential equations to transport data, the flow is deterministic. Deterministic sampling processes
\cite{song2021scorebased, song2020denoising}, have been wildly used with diffusion models for image and video editing and composition as they better preserve semantic content than their stochastic counterparts \cite{mou2024t2i, hertz2023prompttoprompt, Dong2023PromptTI, wang2024cove}.
This property is promising in domain adaptation contexts, where preserving semantics is critical.
Moreover, unlike previously described image translation methods, such as Pix2Pix \cite{isola2017image}, CycleGAN \cite{CycleGAN2017}, or UNSB \cite{kim2023unsb}, FMMs do not rely on adversarial learning to align the endpoint distributions, making them easier to train and less sensitive to hallucinations.

\section{Method}

\begin{figure*}[t]
\vskip 0.2in
\begin{center}
\centerline{\includegraphics[width=0.95\textwidth, trim=35px 0px 0px 0px, clip]{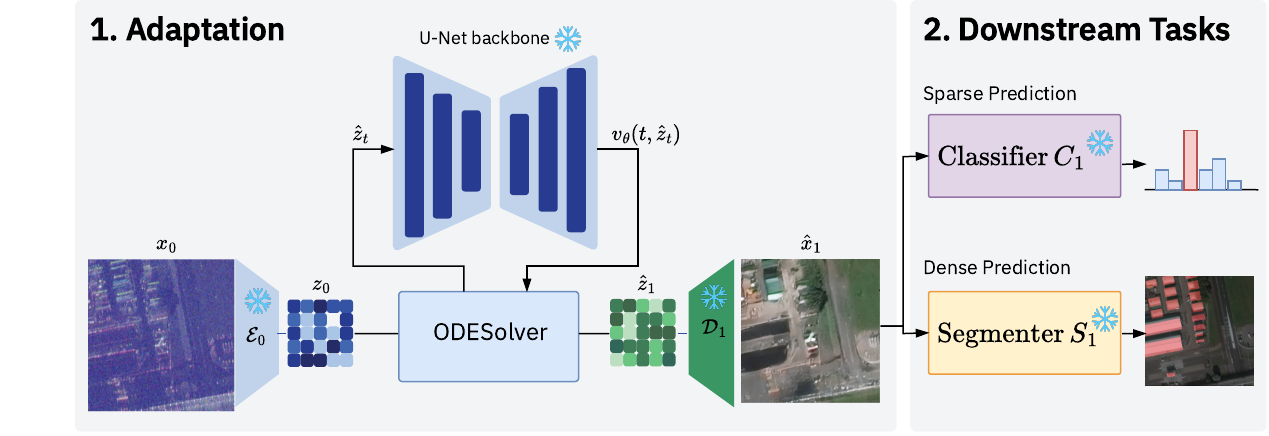}}
\caption{\ours{} offers domain adaptation in image-space, making the adaptation independent of the downstream task and predictive model used. At inference time, we adapt the image $x_0$ into a synthetic image $\hat{x}_1$ by integrating the flow with an ODE solver and the learned velocity $v_\theta$. Then, any predictive model $S_1/C_1$ can directly perform downstream tasks on the transferred images, without fine-tuning.}
\label{fig:main_inference_figure}
\end{center}
\vspace{-10mm}
\end{figure*}

Our goal is to apply an existing classifier or segmenter $S_1$ trained on source domain $D_1$ on a new target domain $D_0$.
We assume that we have access to samples from $D_0$, although we do not know their labels.
To solve this \emph{unsupervised domain adaptation} problem, we introduce \ours to perform domain adaptation in pixel space (see \cref{fig:framework_overview}).

We train a flow matching model to build a bridge between the image distribution $p_0$ of $\mathcal{X}_0$ (target) and $p_1$ of $\mathcal{X}_1$ (source) (see \cref{fig:main_figure}).
Let $\varphi$ be the learned transfer, \ie our mapping from $D_0$ to $D_1$. 
To apply our existing predictive model on data from the target domain, we first map it to the source domain, \ie our model predicts $S_1\left(\hat{x}_1\right)$ (\cref{fig:main_inference_figure}, step 2).
This prediction should be as close as possible to the (unknown) ground truth $y_0$, \ie we want $\varphi$ to preserve the semantic information relevant to the task during transfer.
By transferring the images rather than adapting the predictive model, \ours only depends on the datasets $D_0$ and $D_1$, and neither on the task nor the model $S_1$ (\cref{fig:main_inference_figure}, stage 2). This makes it applicable to a broad panel of tasks, and can benefit from better predictive models without retraining.

\subsection{Training the flow}

\paragraph{Mapping domains} Flow matching models have been used extensively as generative models, mapping a normal distribution to the images' latent distribution, similar to diffusion models \cite{esser2024scaling, liu2024instaflow, kim2025simple, lee2024improving}. However, flow matching can also bridge between arbitrary distributions \cite{lipman2022flow, albergo2023stochastic}.
Following this framework, we leverage a time-dependent flow $\varphi: [0, 1]\times \mathbb{R}^d$ guided by a velocity field $u_t$ describing the trajectories of samples $z$ moving from $p_0$ to $p_1$:
\begin{equation}
\frac{d}{dt}\varphi_t(z) = u_t(\varphi_t(z))
\label{eq:flow}
\end{equation}

The flow $\varphi_t$ results in a transport between $p_0$ and $p_1$ when solving the Ordinary Differential Equation (ODE) defined by \cref{eq:flow} from $t=0$ to $t=1$.
Conversely, solving the same ODE with decreasing times $t=1$ to $t=0$ allows, by construction, to transport $p_1$ to $p_0$.
While the true velocity field $u_t$ is intractable, it is approximated with a neural network $v_\theta(t, z_t)$ trained with a simple regression and simulation-free objective (\cref{fig:main_figure}):
\begin{equation}
    \mathcal{L}_\text{FM}(\theta) = \mathbb{E}_{\substack{z_0, z_1 \sim p(z_0, z_1)\\ t \sim U([0, 1])}} \big\lVert v_\theta(t, z_t) - u_t(z_t\mid z_0, z_1) \big\rVert^2
\label{eq:fmloss}
\end{equation}

where pairs $(z_0, z_1)$ are sampled from the joint distribution $p(z_0, z_1)$ also named \textit{coupling}, that will be detailed later.
From these endpoints, we can build $z_t$ using an interpolant~\cite{albergo2023stochastic}. We use linear interpolants, \ie $z_t = (1-t)z_0 + tz_1$ for which the conditional velocity field  $u_t(z_t\mid z_0, z_1)$  equals $z_1 - z_0$. 
At inference time (\cref{fig:main_inference_figure}, Stage 1), we deploy ODE solvers to solve \cref{eq:flow} by replacing the true velocity $u_t$ with its neural network approximation $v_\theta(t, \cdot)$. This way, we generate the transferred observation $\hat{z}_1$ by integrating the ODE starting from $z_0$ using the mapping $\varphi_t$ following:

\begin{equation}
    \hat{z}_1 = \varphi_{t=1}(z_0) = \text{ODESolver}^{v_\theta}(z_0, 0 \rightarrow 1)
\end{equation}

\paragraph{Latent flow}
With high resolution Earth observation, image dimensions need to be large to include enough spatial context. For example, a $256\times256$ tile covers only $\approx\SI{100}{\meter}\times\SI{100}{\meter}$ at \SI{40}{\centi\meter\per\pixel}.
Because training generative models at a high resolution is compute-intensive, training generative models in the latent space of a \textit{Variational Auto-Encoder} (VAEs) has become a common strategy to improve image generation and accelerate sampling \cite{esser2021taming, rombach2022high, kouzelis2025eq}. We train a flow model in the latent space of a frozen pretrained VAE. Given an image $x$, the VAE's encoder $\mathcal{E}$ compress it into a latent $z = \mathcal{E^*}(x)$ of lower dimensionality. The decoder $\mathcal{D*}$ generates images from latent codes.
Although this VAE was trained exclusively on 3-channel RGB images and not specifically on remote sensing data, the encoder still learns effective representations for such inputs, including non-RGB modalities like SAR. Due to differences in value distributions, the decoder is fine-tuned on SAR and multispectral domains prior to the flow matching training (see \cref{sec:vae_fine-tuning}).
In practice, this means that $p_0$ and $p_1$ represent the distributions of latents $z_0 = \mathcal{E^*}(x_0)$ and $z_1 = \mathcal{E^*}(x_1)$ instead of the images of $\mathcal{X}_0$, $\mathcal{X}_1$ in \cref{eq:flow,eq:fmloss}.

\begin{table*}[ht]
    \centering
    \setlength{\tabcolsep}{4pt}
    \adjustbox{max width=\textwidth}{
    \begin{tabular}{lllrlrc}
        \rowcolor{customgreen!50}
        \textbf{Dataset} & \textbf{Target} & \textbf{Source} & \textbf{Resolution} & \textbf{Task} & \textbf{Size} & \textbf{Alignment}\\
         SpaceNet~6 \cite{shermeyer2020spacenet} & SAR (aerial) & RGB (WorldView-2) & \SI{2}{\meter\per\pixel}& Segmentation & \num{50000} & Strong\\
         Sen1floods11 \cite{Bonafilia_2020_CVPR_Workshops} & SAR (Sentinel-1) & Optical (Sentinel-2) & \SI{10}{\meter\per\pixel} & Segmentation & \num{64512} & Strong\\
         BigEarthNet2 (reBEN) \cite{clasen_reben_2025} & SAR (Sentinel-1) & Optical (Sentinel-2) & \SI{10}{\meter\per\pixel} & Multi-label classification & \num{237871} & Strong\\
         SpaceNet~8 Germany \cite{9857340} & RGB (post-flood) & RGB (pre-flood) & \SI{0.8}{\meter\per\pixel} & Segmentation & \num{5688} & Weak\\
         SpaceNet~8 Louisiana \cite{9857340} & RGB (post-flood) & RGB (pre-flood) & \SI{0.8}{\meter\per\pixel}& Segmentation & \num{17173} & Weak\\
         \bottomrule
    \end{tabular}
    }
    \caption{Datasets used for domain adaptation. We evaluate post-flood to pre-flood adaptation and SAR-to-optical translation scenarios.}
    \label{tab:datasets}
\end{table*}

\subsection{Coupling}\label{section:coupling}

The properties of the transport learnt by the flow are greatly influenced by the choice of the image pairs $(x_0, x_1)$ used to compute the loss function \cref{eq:fmloss}. The most common setup relies on an independent coupling, \ie $(z_0, z_1)$ is sampled uniformly across all possible pairings. Recent works \cite{tong2023improving, liu2022rectified, bortoli2024schrodinger} have introduced couplings inspired by \textit{optimal transport}. However, their optimal transport coupling is defined with the L2-distance between images.
In image space, there is no obvious reason that images with a small pixel-wise Euclidean distance would be semantically similar -- an intuition we show to be true in \cref{seq:couplings}. This contradicts our goal to obtain a transfer that preserves semantics.
Because we leverage flow matching for image translation, \ie conditional generation, we can turn towards a \emph{data-dependent} coupling $p(x_0, x_1) = p(x_1|x_0)p(x_0)$ \cite{albergo2023couplings, somnath2023aligned}. 
We therefore want to build image pairs of a semantically relevant $x_1$ in the $p_1$ distribution, given an image $x_0 \sim p_0$ instead of defining a new ad hoc joint distribution. Finally, to sample latents from $p(z_0, z_1)$, we first sample from the image coupling $p(x_0, x_1)$ and then encode the images. 

\subsection{Alignment}\label{section:alignment}
\begin{figure}[h!]
    \hspace*{-10px} 
    \centering
    \input{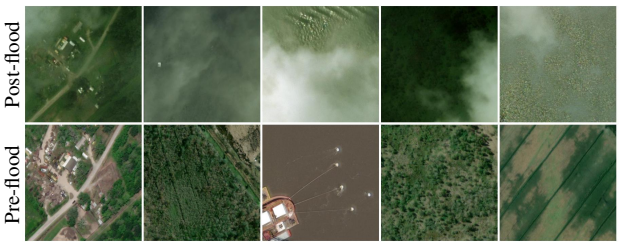}
    \caption{Weakly-aligned image pairs from the SpaceNet 8 dataset, affected by cloud coverage and natural disasters. Each column:  top=post-flooding imagery; bottom=pre-event imagery.}
    \label{fig:weakly_aligned_dataset}
\end{figure}

We aim at building pairs of images $(x_0, x_1)$ that are semantically close. Because remote sensing data is geospatial, the coordinate metadata are available for each image. Thus, we consider spatially aligned datasets, which is possible to construct in most real-world applications, and leave geographical domain adaptation for future work.
While coregistration provides pairs of images that have a common location, it does not ensure that the semantic information is shared between the two images. Thus, we distinguish between:

\begin{itemize}
    \item \textbf{Strong semantic alignment}: the two images $x_0$ and $x_1$ share the same semantics, \ie  $y_0 = y_1$. This is the ideal scenario, though impractical, as it needs synchronized acquisitions, or at least images of the same areas in a short timeframe, such that no significant semantic changes have occurred. This can be typically used to address sensor shift, \eg, SAR to optical translation.
    
    \item \textbf{Weak semantic alignment}: the two images $x_0$ and $x_1$ partially share their semantics, \ie the ground truths are similar $y_0 \approx y_1$. 
    For example, these may be acquisitions of the same geographical area but captured on different dates. As shown in \cref{fig:weakly_aligned_dataset},changes in semantics may be due to the construction of buildings between the two acquisitions, harvested crops, cloud coverage, natural disasters such as floods or fires, deforestation, moving objects, or any other event that can shift semantics.
    Note that $x_1$ in the dataset is not an accurate representation of the transferred $x_0$ because of the changes. Yet, in the absence of labels, this pair $(x_0, x_1)$ is the best available coupling. We then assume that the averaging of velocities in \cref{eq:fmloss} is robust to moderate semantic changes and preserves the main transfer components between $p_0$ and $p_1$.
    This can be used to address temporal shift, \eg seasonal variations, and before/after an extreme event.
\end{itemize} 

\section{Experimental setup}

\begin{table*}[t]
\centering
\setlength\tabcolsep{8pt}
\centering
\resizebox{0.99\linewidth}{!}{%
    \begin{tabular}{l cccc|cccc|cccc}
        \toprule
                 \rowcolor{customgreen!50} 
Datasets& \multicolumn{4}{c}{SpaceNet 8} & \multicolumn{4}{c}{SpaceNet 8 Germany}& \multicolumn{4}{c}{SpaceNet 8
Louisiana}\\ 
             & \multicolumn{4}{c}{Post-flood $\rightarrow$ Pre-flood} & \multicolumn{4}{c}{Post-flood $\rightarrow$ Pre-flood}& \multicolumn{4}{c}{Post-flood $\rightarrow$ Pre-flood}\\
                 & mIoU $\uparrow$& mAcc $\uparrow$& FID $\downarrow$& LPIPS $\downarrow$& mIoU $\uparrow$& Acc $\uparrow$& FID $\downarrow$& LPIPS $\downarrow$& mIoU $\uparrow$& mAcc $\uparrow$& FID $\downarrow$& LPIPS $\downarrow$ \\
            No adaptation   & 40.05& 42.40& 75.62& 63.66& 37.09& 39.08& 89.54& 63.27& 36.51& 38.85& 96.60&63.80\\
            Upper bound    & 63.10& 72.09& 00.00& 00.00& 55.27& 66.77& 00.00& 00.00& 66.91& 75.97& 00.00&00.00\\ 
 \midrule
            Pix2Pix& 34.73& 36.08& 98.22& \underline{50.95}& 32.92& 34.25& 98.38& \underline{55.75}& 38.79& 40.86& 92.23&\underline{47.05}\\ %
            CycleGAN& \underline{40.70}& \underline{43.35}& \textbf{54.31}& 55.70& \underline{39.35}& \underline{41.79}& \textbf{62.80}& 59.46& \underline{42.39}& \underline{45.14}& \textbf{52.80}&52.92\\ %
            UNSB& 39.35& 42.67& 68.30& 55.35& 38.25& 40.62& \underline{66.62}& 56.84& 40.67& 43.87& 73.72&53.04\\
            Diffusion Bridge& 37.50 & 39.36 & 115.70 & 53.13 & 33.91 & 35.27 & 177.23 & 58.53 & 39.05 &  41.37 & 105.27 & 51.25\\
            StegoGAN& 38.62& 40.58& 66.61& 58.07& 36.74& 38.78& 90.42& 63.50& 40.14& 42.29& 68.56&54.58\\ %
 \midrule       
            \rowcolor{customblue!50} \textbf{FlowEO}& \textbf{44.65}& \textbf{48.79}& \underline{60.32}& \textbf{45.50}& \textbf{41.27}& \textbf{45.29}& 82.74& \textbf{53.63}& \textbf{47.19}& \textbf{52.30}&  \underline{59.65}& \textbf{41.95}\\
\bottomrule
    \end{tabular}
    }
    \caption{Quantitative results on domain adaptation for weakly aligned datasets. We report both segmentation (mIoU, mAcc) and image quality metrics (FID, LPIPS) for SpaceNet 8 and its geographic subsets. \ours{} transports images while preserving its semantics, achieving significant segmentation performance improvements in domain adaptation setting: \num{44.65} \vs \num{40.05} mIoU on SpaceNet 8. It also outperforms the second-best model -- CycleGAN -- on segmentation accuracy after transfer.}
\label{tab:da_weakly_results}
\end{table*}

\begin{table*}[ht]
\centering
\setlength\tabcolsep{8pt}
\centering
\resizebox{0.99\linewidth}{!}{%
    \begin{tabular}{l cccc|cccc|cccccc}
        \toprule
             \rowcolor{customgreen!50}Datasets& \multicolumn{4}{c}{Sen1Floods1} & \multicolumn{4}{c}{SpaceNet 6}& \multicolumn{6}{c}{ReBEN}\\ 
             & \multicolumn{4}{c}{SAR $\rightarrow$ Optical} & \multicolumn{4}{c}{SAR $\rightarrow$ RGB}& \multicolumn{6}{c}{SAR $\rightarrow$ Optical}\\
                 & mIoU& mAcc& FID& LPIPS & mIoU& mAcc& FID& LPIPS& $\text{AP}^\mu$& $\text{AP}^\text{M}$& $\text{F1}^\mu$&$\text{F1}^\text{M}$& FID&LPIPS  \\
   No adaptation   & 06.22& 49.72& 297.22& 84.84& 31.94& 41.01& 275.05& 79.48& 17.46& 17.43& 02.31&01.31& 339.36&85.99\\
    Upper bound    & 55.14& 71.28& 00.00& 00.00& 84.94& 90.74& 00.00& 00.00& 79.26& 65.28& 74.28&62.84& 00.00&00.00\\ 
 \midrule
            Pix2Pix& \underline{51.50}& \underline{62.31}& \underline{20.64}& \underline{31.33}& \underline{56.48}& \underline{63.43}& 130.42& \underline{41.89}& \textbf{41.09}& \underline{27.88}& \textbf{43.93}& \textbf{25.79}& \textbf{62.84}&\underline{17.56}\\ %
            CycleGAN& 42.12& 48.47& 20.97& 36.35& 50.01& 55.85& 132.75& 50.72& 26.09& 19.79& 26.93&15.75& 81.54&19.67\\ %
            UNSB& 42.69& 48.85& 23.01& 35.01& 52.43& 61.04& \textbf{72.48}& 45.81& 25.61& 20.71& 29.52&19.45& 113.73&35.64\\
            Diffusion Bridge& 42.41 & 50.31 & 18.71 & 39.93 & 51.22 & 58.37 & 94.15 & 46.37 & 18.44 & 15.79 & 24.43 & 05.80 & 80.97 & 20.74\\
             StegoGAN& 43.37& 49.75& 41.06& 31.87& 44.87& 50.02& 306.50& 56.62& 26.13& 22.16& 29.49&20.28& 81.15&22.32\\ %
 \midrule       
             \rowcolor{customblue!50}\textbf{FlowEO}& \textbf{54.92}&\textbf{ 69.04}& \textbf{12.96}& \textbf{29.21}& \textbf{65.07}& \textbf{72.33}& \underline{94.02}& \textbf{39.96}& \underline{37.16}& \textbf{32.14}&  \underline{36.04} & \underline{25.72}& \underline{75.80}&\textbf{15.51}\\
\bottomrule
    \end{tabular}
    }
    \caption{Quantitative results on domain adaptation for strongly aligned datasets. We report both segmentation (mIoU, mAcc) or classification (AP/F1) and image quality metrics (FID, LPIPS). \ours preserves achieves the best UDA segmentation performances, and on-par classification performances with Pix2Pix.}
\label{tab:da_strong_results}
\end{table*}

\subsection{Datasets}

We evaluate \ours for domain adaptation on three segmentation and one classification datasets, listed in \cref{tab:datasets}: SpaceNet~6 \cite{shermeyer2020spacenet}, Sen1floods11 \cite{Bonafilia_2020_CVPR_Workshops}, BigEarthNet2 (reBEN) \cite{clasen_reben_2025} and SpaceNet~8 \cite{9857340}, split into Germany and Louisiana.
These datasets are paired, \ie have multiple acquisitions for the same area, allowing us to train image translation models with data dependent couplings. SpaceNet~8 contains before/after images of flood event, semantic differences exist in the images, making it ``weakly aligned''. The others pair images from close dates, resulting in a ``strong'' alignment. We build 3-channels $256\times256$ images using RGB for color images, bands $[4,3,2]$ for Sentinel-2, VV/HH/VH polarizations for SAR images from SpaceNet~6 and reBEN, and VV/VH/VH for Sen1Floods11. See \cref{sup:dataset} for details.

\paragraph{Downstream models}
We use the DeepLabv3+ architecture~\cite{chen_encoder-decoder_2018} for semantic segmentation with a ResNet-34 backbone, ImageNet initialization, a batch size 512, and a learning rate of \num{0.001} with one-cycle cosine schedule. For classification, we follow the reBEN implementation \cite{clasen_reben_2025} and train a ResNet-50 with ImageNet initialization for \num{100000} training steps with a batch size of 512 and
a linear-warmup-cosine-annealing learning rate of 0.001.
These models are trained once and used to evaluate all image translation methods.

\subsection{Model comparison}

\paragraph{Baselines}
We compare our \ours model against several commonly used image translation baselines: Pix2Pix~\cite{isola2017image}, CycleGAN~\cite{CycleGAN2017}, StegoGAN~\cite{wu2024stegogan}, Diffusion Bridge~\cite{bortoli2024schrodinger},  and Unpaired Neural Schrödinger Bridges (UNSB)~\cite{kim2023unsb}. For a fair comparison, we use data-dependent coupling for all methods, even those that could be trained using independent couplings (CycleGAN, StegoGAN, and UNSB), and train all models for \num{200000} steps. We follow official implementations for hyperparameters (cf. \cref{sup:hyperparameters}). Except CycleGAN, baselines are non-symmetric, thus we train two separate models from domain $\mathcal{X}_0$ to $\mathcal{X}_1$ and then from $\mathcal{X}_1$ to $\mathcal{X}_0$ when needed.

\paragraph{Hyperparameters}
We train our flow matching in the pretrained space of the VAE from Stable Diffusion 3 \cite{esser2024scaling}. More precisely, we use a distilled model that is smaller and more compute efficient \cite{bohan_madebyollin_2025}.
The flow is therefore performed on the latent codes of dimensions $16\times32\times32$.
Because flow matching is symmetrical, the same model can be used to transfer from $\mathcal{X}_0$ to $\mathcal{X}_1$ and vice versa, while baselines require two models.
We use the classical U-Net backbone to train the flow \cite{song2021scorebased} with 120 million parameters.
The flow is trained for \num{200000} steps using gradient clipping and exponential moving average. We use a learning rate of \num{1e-4} with \num{1000} steps of linear warmup and a batch size of 256.
At inference time, we integrate the flow from \cref{eq:flow} with \num{50} steps of the Euler ODE sampler. We use the sigmoid time-scheduler introduced in \cite{kim2025simple}
to focus on the times that are close to the image spaces. See \cref{supp:sigmoid} for more insights and ablation studies about sampler design and \cref{supp:memory} for inference time and memory footprints.

\subsection{Evaluation}

\paragraph{Prediction metrics} Because Earth observation tasks are often dense predictions, we focus on domain adaptation for semantic segmentation. For all methods, we first transfer the images from the test set of each dataset using the image translation model and then apply the same pretrained segmenter to obtain the semantic masks. We then compute \textit{mean Intersection over Union} (mIoU) and \textit{mean Accuracy} (mAcc) between the prediction on the transferred image $\hat{x}_0 = \varphi(x_1)$ and the ground truth mask $y_0$, \emph{that is only available for evaluation purposes}. For reBEN, we use the standard multi-label classification metrics: \textit{Average Precision} (AP) and \textit{$F_1$-score} ($F_1$), both micro and macro, \ie $\text{AP}^\mu$, $F_1^\mu$, $\text{AP}^M$, $F_1^M$.

\begin{figure*}[ht]
\centering
\input{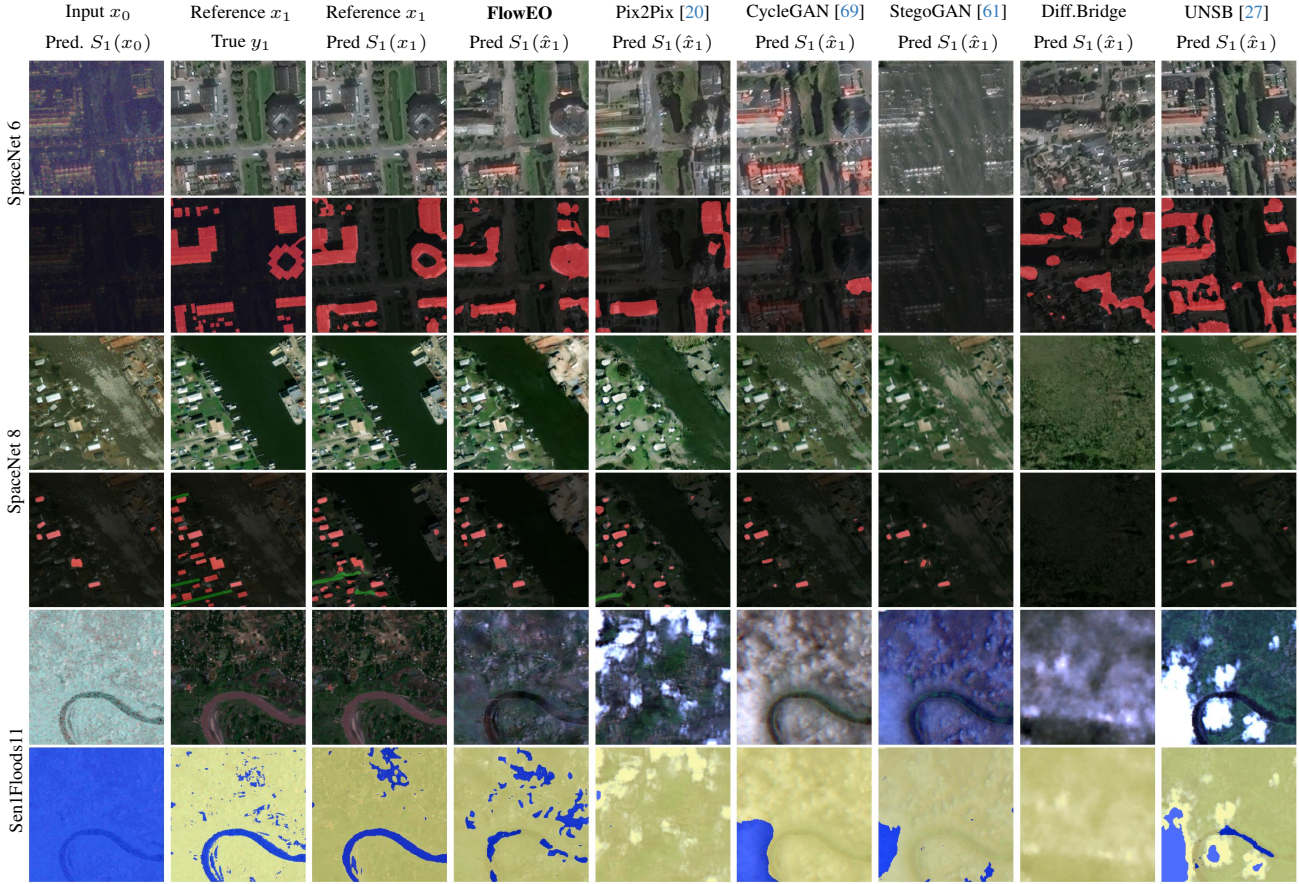}
\caption{Qualitative comparison of domain adaptation methods on segmentation datasets. The first column represents the input image $x_0$, the second and third depict the weakly or strongly aligned $x_1$, and the others display the images generated by the different methods. Below each image, we provide the corresponding prediction from the segmentation model $S_1$ or the true segmentation mask $y_1$ for the reference image (third column). \ours{} outperforms other methods in both semantic preservation and image quality.}
\label{fig:viuals}
\end{figure*}

\paragraph{Image quality} While it is not our main goal, we also evaluate the perceptual quality of the generated images as visual artifacts can hinder downstream performances and interpretability. We compute both the \textit{Frechet Inception Distance} (FID) \cite{fid} and \textit{LPIPS} \cite{zhang2018perceptual} similarity between the transferred images $\hat{x}_0$ and source images $x_1$.
Although commonly used, note that these are initially designed for natural images and not remote sensing imagery \cite{jayasumana2024rethinking}. In addition, the size of our test sets is under the recommended size to compute FID. Despite the noise this might introduce, these metrics remain useful proxies to assess broad tendencies regarding the perceptual qualities of transferred images.

\section{Results}

\subsection{Main results}

We present domain adaptation and image quality metrics obtained by the compared image translation methods in \cref{tab:da_weakly_results} (weakly aligned) and \cref{tab:da_strong_results} (strongly aligned). In addition to the results obtained by state-of-the-art models and \ours{}, we include two comparison points:

\begin{itemize}
    \item \textbf{No adaptation}: classification/segmentation metrics of the pretrained model applied directly on the non-transferred target data, \ie the performance of $S_1$ on $x_0$. This represents a lower bound of the expected performance. Image quality metrics (FID and LPIPS) are computed directly between images from $\mathcal{D}_0$ and $\mathcal{D}_1$ and show an estimate of how far away the two image distributions are.
    \item \textbf{Upper bound}: classification/segmentation metrics of the pretrained model on its source domain, \ie the performance of $S_1$ on $x_1$. This represents the performance of an ideal semantic-preserving transfer from $p_0$ to $p_1$, for which $S_1$ is as accurate on $x_0$ transferred as on $x_1$.
\end{itemize}

\paragraph{Semantic preservation}
\ours{} consistently demonstrates superior semantic preservation compared to existing image translation models. It ranks first in the weakly aligned setting (\cref{tab:da_weakly_results}), significantly outperforming both the second-best state-of-the-art transfer method (+4 mIoU points compared to CycleGAN) and the no-transfer baseline by a large margin (\num{44.65} mIoU vs. \num{40.05} mIoU) on SpaceNet~8.
Domain adaptation for pre/post-flood imagery is a particularly challenging task considering the significant changes that impact the images, as shown in \cref{fig:viuals}.
Note that only \ours and CycleGAN successfully increase segmentation performance over the no-adaptation baseline.
\ours consistently achieves the highest mIoU and mean accuracy across both regions (Germany and Louisiana), demonstrating its effectiveness in handling real-world geographic variations, even when trained on smaller datasets ($<$\num{10000} samples).

For strongly aligned datasets (\cref{tab:da_strong_results}), \ours{} achieves the best segmentation metrics on both Sen1Floods11 and SpaceNet 6 datasets with respectively +\num{3.42} and +\num{8.59} in mIoU compared to the second best transfer models.
Somewhat surprisingly, Pix2Pix constitutes a strong baseline for paired image translation and achieves the second-best performance in this setting despite being the oldest model evaluated.
On the ReBEN multi-label classification dataset, the flow model and Pix2Pix perform competitively, trading first and second places depending on the metric considered.
Despite their training with data-dependent coupling, adversarial-based methods struggle to offer semantic-preserving transport. This suggests that adversarial objectives may be unaligned with semantic preservation by hallucinating new instances \eg clouds, that can reduce segmentation performances (as shown for Sen1Floods11 in \cref{fig:viuals}, fifth row)

\paragraph{Transferred image quality}
In addition to better preserving the semantics, \ours generates consistent high-quality images.
It ranks first in LPIPS and first or second in FID on all datasets, both in weakly aligned RGB$\rightarrow$RGB transfer on SpaceNet~8 (\cref{tab:da_weakly_results}) and strongly aligned SAR-to-optical translation (SpaceNet~6, Sen1Floods11 and reBEN in \cref{tab:da_strong_results}). 
Unlike previous methods, \ours{} does not rely on adversarial loss functions explicitly designed to enhance perceptual quality. Despite that, generated images remain of high quality and do not show hallucinations commonly attributed to adversarial training.
This trend holds for both weakly and strongly aligned datasets. In particular, we observe that \ours{} learns complex texture transfer on the post-to-pre-disaster scenario, correctly mapping turbulent and murky flood water to the usual river state (\cref{fig:viuals}, third row).

\subsection{Impact of the coupling}
\label{seq:couplings}

We report in \cref{tab:coupling_trained} ablation results on SpaceNet~6 and SpaceNet~8 domain adaptation and image generation using the three couplings: independent, minibatch-OT \cite{tong2023improving}, and data-dependent.
Minibatch-OT coupling outperforms the independent coupling on the two datasets in segmentation accuracy after domain adaptation and image quality.
Yet, data-dependent coupling outperforms them by a large margin for domain adaptation (+17\% mIoU on SpaceNet~6, +7\% mIoU on SpaceNet~8) and image quality (30\% decrease in FID on both datasets).
This is expected since OT pairs images based on Euclidean distance in pixel space, which is irrelevant to semantics \eg in SpaceNet~6 where it compares SAR and RGB modalities.
Yet, OT is also far behind the data-dependent coupling in the favourable case of RGB to RGB transport on SpaceNet~8.  
This confirms the importance of data-dependent coupling -- thus dataset alignment -- to preserve semantic information during flow-based transfer, and motivates our focus on aligned datasets, even weakly.

\begin{table}[t]
\centering
\setlength\tabcolsep{2pt}
    \resizebox{\linewidth}{!}{%
\begin{tabular}{l c c c c}
  \rowcolor{customgreen!50}\multicolumn{5}{c}{SpaceNet 8}\\ %
         &\multicolumn{4}{c}{Post-flood $\rightarrow$ Pre-flood}\\
  Coupling & mIoU $\uparrow$& mAcc $\uparrow$& FID$\downarrow$& LPIPS $\downarrow$ \\
  Independent $p(x_0)p(x_1)$&35.59& 37.41& 94.23& 66.62\\
  Minibatch-OT $\pi(x_0, x_1)$& 37.26& 39.28& 84.44& 63.93\\
  Data-dependent $p(x_1|x_0)p(x_0)$& \textbf{44.65}& \textbf{48.79}& \textbf{60.32}& \textbf{45.50}\\
  & \multicolumn{4}{c}{Pre-flood/Post-flood}\\
  Coupling & mIoU $\uparrow$& mAcc $\uparrow$& FID $\downarrow$& LPIPS $\downarrow$\\
  Independent $p(x_0)p(x_1)$ & 35.60& 37.80& 80.26& 67.88\\
  Minibatch-OT $\pi(x_0, x_1)$& 36.21& 39.28& 73.26& 65.22\\
  Data-dependent $p(x_1|x_0)p(x_0)$& \textbf{44.87}& \textbf{53.76}& \textbf{50.88}& \textbf{52.81}\\
   \rowcolor{customgreen!50}\multicolumn{5}{c}{SpaceNet 6}\\ %
             &\multicolumn{4}{c}{SAR $\rightarrow$ RGB}\\
  Coupling & mIoU $\uparrow$& mAcc $\uparrow$& FID$\downarrow$& LPIPS $\downarrow$ \\
  Independent $p(x_0)p(x_1)$&45.25 & 50.75& 145.02& 65.94\\
  Minibatch-OT $\pi(x_0, x_1)$& 48.48& 55.03& 125.82& 58.34\\
  Data-dependent $p(x_1|x_0)p(x_0)$& \textbf{65.07}& \textbf{72.33}& \textbf{94.02}& \textbf{39.98}\\
  & \multicolumn{4}{c}{RGB $\rightarrow$ SAR}\\
  Coupling & mIoU $\uparrow$& mAcc $\uparrow$& FID $\downarrow$& LPIPS $\downarrow$\\
  Independent $p(x_0)p(x_1)$ & 45.74& 50.85& 105.47& 64.69\\
  Minibatch-OT $\pi(x_0, x_1)$& 47.25& 52.65& 91.74& 60.24\\
  Data-dependent $p(x_1|x_0)p(x_0)$& \textbf{55.36}& \textbf{61.53}& \textbf{36.86}& \textbf{51.66}\\
    \bottomrule
    \end{tabular}
}
    \caption{Impact of coupling on generation quality and semantic preservation during transfer. The OT-based coupling $\pi(x_0, x_1)$ fails to match the performance of the data-dependent coupling $p(x_1|x_0)p(x_0)$, although it outperforms the independent coupling.}
\label{tab:coupling_trained}
\vspace{-2mm}
\end{table}

\section{Conclusion}

We introduce \ours{}, a flow matching-based framework for unsupervised domain adaptation in Earth Observation. By learning a semantically consistent mapping between source and target distributions, \ours{} consistently outperforms existing image translation methods for domain adaptation in five segmentation and classification tasks across multiple challenging scenarios ranging from post-disaster monitoring to SAR-to-Optical translation, while achieving on-par or better image generation quality. \ours opens the door to generic unsupervised domain adaptation with possible extensions to semantic-based couplings based on image similarity or image metadata embeddings to fare with unpaired image translation scenarios in Earth observation.

\section*{Acknowledgement}

This work was conducted as part of the research project \href{https://mage.science/}{\textsc{MAGE}} (ANR-22-CE23-0010) funded by the \emph{Agence
Nationale de la Recherche}. This work was granted access to the HPC resources of IDRIS under the allocation AD011014327R2 made by GENCI.

{\small
\bibliographystyle{ieee_fullname}
\bibliography{main}
}

\newpage
\appendix
\section{Details and ablations}

\subsection{Couplings}

\begin{figure}[h!]
    \centering
    \begin{minipage}[b]{0.32\columnwidth}
        \centering
        \includegraphics[width=\textwidth]{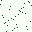}
    \end{minipage}
    \hfill
    \begin{minipage}[b]{0.32\columnwidth}
        \centering
        \includegraphics[width=\textwidth]{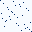}
    \end{minipage}
    \hfill
    \begin{minipage}[b]{0.32\columnwidth}
        \centering
        \includegraphics[width=\textwidth]{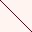}
    \end{minipage}
    \caption{Comparison between the pairing matrices generated with the different couplings for a batch on SpaceNet 8, from left to right: independent coupling $p(x_0)p(x_1)$, OT-coupling $\pi(x_0, x_1)$, data-dependent coupling $p(x_1\mid x_0)p(x_0)$.}
    \label{fig:couplings}
\end{figure}

The choice of the coupling has been of prime importance to improve generation capabilities for flow matching models \cite{tong2023improving, liu2022rectified, bortoli2024schrodinger}. \Cref{fig:couplings} shows the pairing matrices $M$ obtained with each coupling \ie $M_{ij}=1$ iff latents $x_0^i$ and $x_1^j$ are paired. The training batches are built by stacking strongly or weakly aligned $x_0$ and $x_1$ images in order. Because the data-dependent coupling matches $x_0^i$ with $x_1^i$, its pairing matrix is diagonal. We observe that the optimal transport-based coupling (left) is poorly aligned with the data-dependent coupling (center), suggesting that semantic information matching cannot be solely recovered through optimal transport.

\begin{figure*}[ht]
\centering
\input{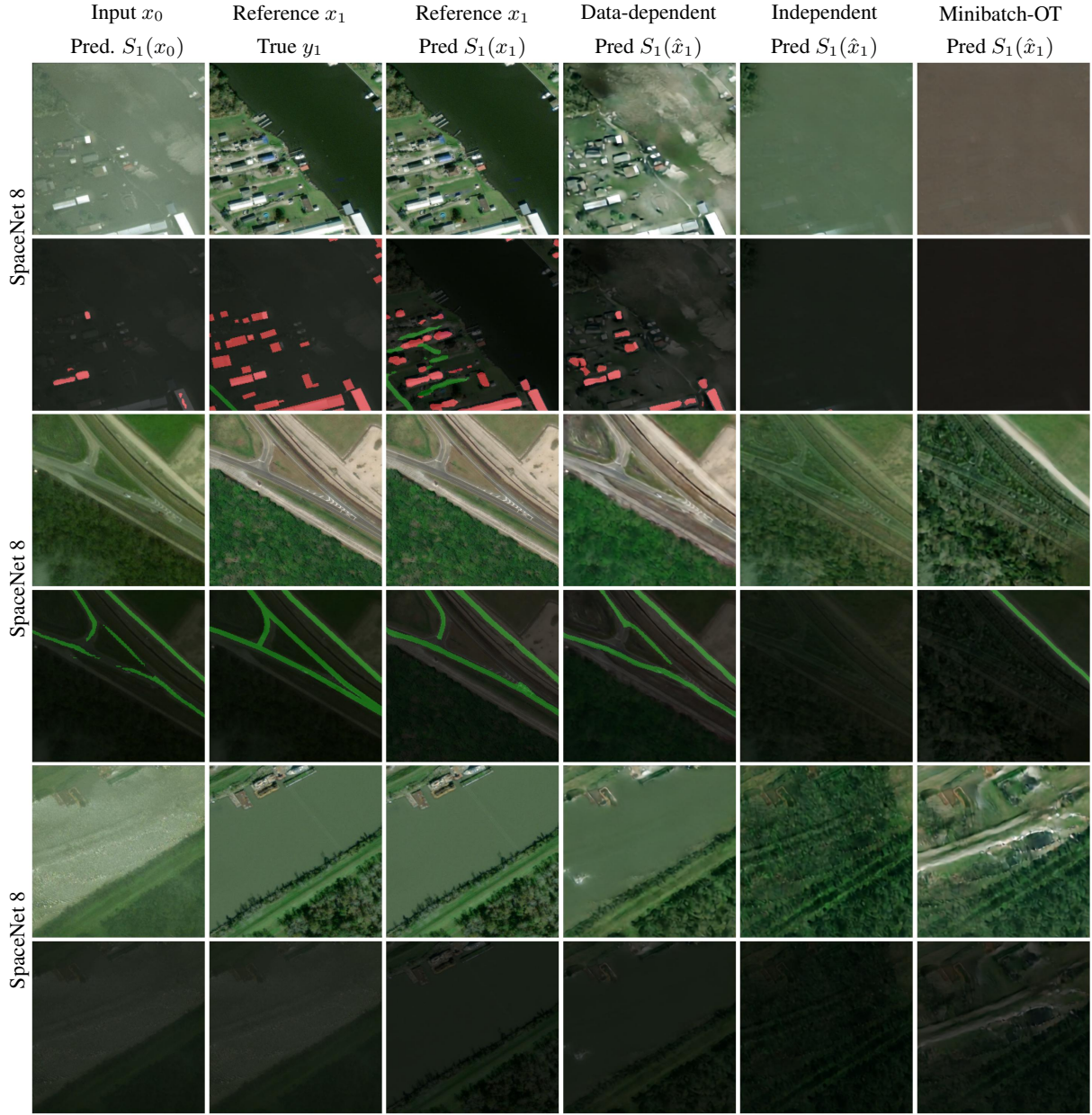}
\caption{Impact of the training coupling $p(x_0, x_1)$ on preserving semantic information during image translation. \ours employs \textit{data-dependent coupling} $p(x_1|x_0)p(x_0)$, which outperforms both \textit{minibatch-OT} coupling $\pi(x_0, x_1)$ and \textit{independent} coupling $p(x_0, x_1)$.}
\label{fig:visual_coupling}
\end{figure*}

In addition, we provide visual ablation results in \cref{fig:visual_coupling}, which illustrate the necessity to use data-dependent couplings to train \ours.

\subsection{VAE finetuning}\label{sup:vae_finetuning}

\begin{table}[h]
\centering
\resizebox{\linewidth}{!}{%
\begin{tabular}{c l c c c c}
\rowcolor{customgreen!50} & \multicolumn{5}{c}{SpaceNet 8 Post-flood $\rightarrow$ Pre-flood}\\
\multirow{3}{*}{\rotatebox{90}{RGB}} & & mIoU $\uparrow$ & mAcc $\uparrow$ & FID $\downarrow$ &LPIPS $\downarrow$ \\
& Base & \textbf{44.65} & \textbf{48.79} & \textbf{60.32} & \textbf{45.50} \\
& Finetuned & 44.33 & 48.71 & 81.75 & 51.64 \\
\rowcolor{customgreen!50} & \multicolumn{5}{c}{SpaceNet 6 SAR $\rightarrow$ RGB}\\
\multirow{3}{*}{\rotatebox{90}{RGB}} & & mIoU $\uparrow$ & mAcc $\uparrow$ & FID $\downarrow$ &LPIPS $\downarrow$ \\
& Base & \textbf{65.07} & \textbf{72.33} & \textbf{94.02} & \textbf{39.96} \\
& Finetuned & 64.63 & 72.17 & 111.66 & 42.77 \\
\rowcolor{customgreen!50} & \multicolumn{5}{c}{Sen1Floods11 SAR $\rightarrow$ Optical}\\
\multirow{3}{*}{\rotatebox{90}{S2}} & & mIoU $\uparrow$ & mAcc $\uparrow$ & FID $\downarrow$ &LPIPS $\downarrow$ \\
& Base & 51.45 & 57.63 & 24.33 & 29.22 \\
& Finetuned & \textbf{54.92} & \textbf{69.04} & \textbf{12.96} & \textbf{29.21} \\
\rowcolor{customgreen!50} & \multicolumn{5}{c}{ReBEN SAR $\rightarrow$ Optical}\\
\multirow{3}{*}{\rotatebox{90}{S2}} & & $\text{AP}^\text{M}$& $\text{F1}^\text{M}$& FID$\downarrow$&LPIPS $\downarrow$  \\
& Base & 27.02 & 15.97 & 168.85 & 16.88 \\
& Finetuned & \textbf{32.14} & \textbf{25.72} & \textbf{75.80} & \textbf{15.51} \\
\bottomrule
\end{tabular}%
}
\caption{Impact of VAE fine-tuning on domain adaptation performance and transferred image quality. Fine-tuning is beneficial for Sentinel-2 imagery but not for classical RGB images.}
\label{tab:vae_finetuning}
\end{table}

\subsubsection{Implementation details}

We use a distilled version of the VAE from StableDiffusion~3 \cite{esser2024scaling} to speed up training and inference. The encoder is trained to reconstruct the latents produced by the original encoder to preserve the latent space structure of the full model. As shown in the main paper, our experiments show that the reconstructions $\mathcal{D}(\mathcal{E}(x))$ of Sentinel-2 images are of poor quality because the range and distribution of multispectral images deviates from the pretraining dataset used for Stable Diffusion. For the reBEN and Sen1Floods11 datasets that use Sentinel-2 as source data, we finetune the decoder of the distilled VAE on each dataset for \num{5000} iterations with a learning rate of $10^{-4}$, \num{250} warmup steps, and cosine decay learning rate scheduler. The decoder remains frozen when training the flow. The remaining datasets use the original pretrained decoder.

\subsubsection{Impact of VAE fine-tuning}
\label{sec:vae_fine-tuning}

\paragraph{Reconstruction} SD VAE reconstruction error is higher on non-RGB imagery, VAE finetuning improves reconstruction RMSEs \num{237.04} \vs\num{357.91} and \num{0.058} \vs\num{3.760} on respectively reBEN S2 and SpaceNet-6 SAR.
This is unnecessary for RGB and can be slightly detrimental. S2 images are normalized from [0;10000] to [-1;+1] via band-wise min-max normalization.

\paragraph{Generation} We report in \cref{tab:vae_finetuning} metrics for flow models trained with and without a fine-tuned VAE decoder.
We observe that fine-tuning the VAE decoder prior to learning the flow matching has a positive impact when the final domain differs from usual RGB imagery. Indeed, fine-tuning the decoder is beneficial for Sen11Floods11 and ReBEN, for which the images are transferred in the Sentinel-2 color bands. Because Sentinel-2 imagery uses the [0, \num{10000}] range instead of the usual [0, 255], the pretrained decoder is less effective, which reflects in image quality. Yet, on SpaceNet~6 and 8, which use both standard RGB images, there is no advantage of fine-tuning the decoder. It is even detrimental, as we hypothesize that the decoder overfits to the small training set, compared to the original dataset used for StableDiffusion.

\subsection{Sampling schedule}
\label{supp:sigmoid}

\begin{figure}[h]
    \centering
    \includegraphics[width=\linewidth, trim=8px 0px 0px 0px, clip]{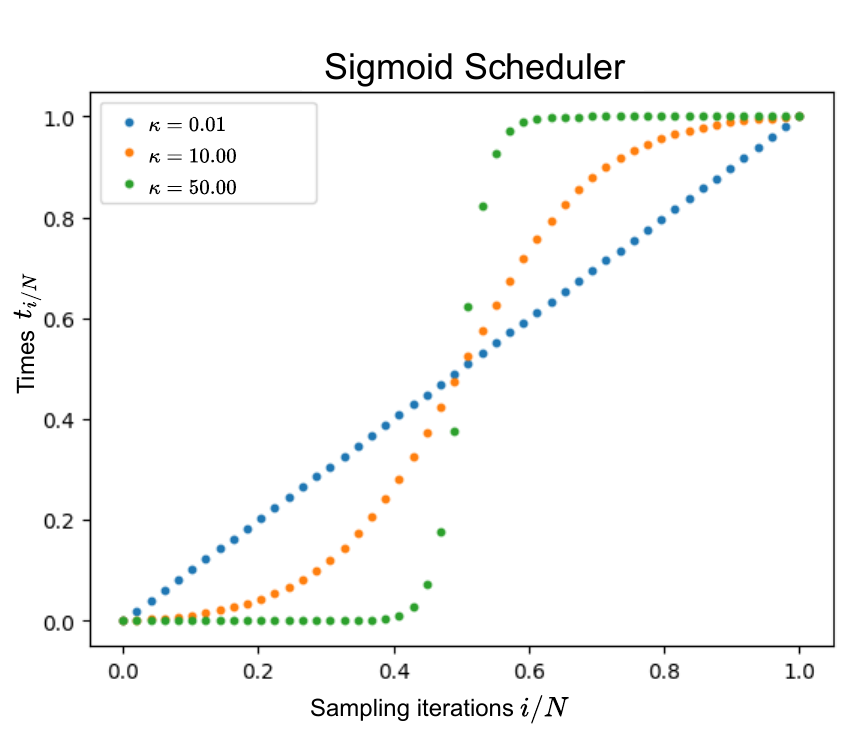} 
    \caption{Sigmoid time discretization, allocating more sampling steps near the endpoints ($t\approx0$ and $t\approx1$).}
    \label{fig:sampling_schedule}
\end{figure}

The choice of time discretization and inference-time sampling strategy plays a crucial role in improving the performance of diffusion models \cite{dpmsolver, lu2023dpmsolver, karras2022elucidating}. Recently, \cite{kim2025simple} introduced a sigmoid time-scheduler tailored for flow matching models (see \cref{eq:sigmoid_schedule}). This scheduler is parametrized by $\kappa$ which controls the distribution of sampling steps across time. Higher values of $\kappa$ concentrate computational effort near the endpoints ($t\approx0$ and $t\approx1$), whereas $\kappa \rightarrow 0$ corresponds to the linear time schedule (see \Cref{fig:sampling_schedule}).
\begin{equation}
    \Bigg\{t_i = \frac{\text{sig}\Big(\kappa\Big(\frac{i}{N}-0.5\Big)\Big)- \text{sig}\Big(-\frac{\kappa}{2}\Big)}{\text{sig}\Big(\frac{\kappa}{2}\Big)-\text{sig}\Big(-\frac{\kappa}{2}\Big)}:i=0, ..., N\Bigg\}
\label{eq:sigmoid_schedule}
\end{equation}

Despite originally designed for generative modeling with flow matching models, \ie mapping a Gaussian prior distribution to the data distribution, this time scheduling is well-motivated in our setting where increasing the number of sampling steps near the data distributions $p_0$ and $p_1$ is beneficial. \cref{tab:scheduler_metrics} presents a comparison between sigmoid and linear time discretization, demonstrating consistent improvements in segmentation metrics across all datasets and for all numbers of inference steps. Image quality metrics exhibit only marginal improvements and, in some cases—such as on the Sen1Floods11 dataset—even show slight deterioration. Nevertheless, the performance gains in segmentation metrics from using a sigmoid rather than a linear schedule diminish as the number of inference steps increases. Also observe that more sampling steps might not be beneficial for domain adaptation. On the two datasets used for validation, 25 sampling steps tends to perform on-par or better than 50 and 100 steps. We attribute this to slightly better preservations of semantics with a low number of steps, which reduce small but accumulating errors in the Euler integration. In practice, we set $\kappa=10$ and use \num{50} sampling steps for all experiments.
\begin{table}[h]
\centering
    \resizebox{\linewidth}{!}{%
\begin{tabular}{l c c c c} %
\rowcolor{customgreen!50} \multicolumn{5}{c}{Sen1Floods11 SAR $\rightarrow$ Optical}\\
  & mIoU $\uparrow$& mAcc $\uparrow$& FID$\downarrow$& LPIPS $\downarrow$ \\
 25 Sampling Steps& & & &\\
  Linear&54.60& 72.22& \textbf{13.99}& \textbf{28.91}\\
  Sigmoid $\kappa=10$ & \textbf{55.05}& \textbf{72.50}& 14.38& 29.02\\
  \midrule
  50 Sampling Steps& & & & \\
  Linear& 54.26&71.79&\textbf{13.06}&\textbf{28.86}\\
  Sigmoid $\kappa=10$ & \textbf{54.46}& \textbf{71.94}& 13.46& 28.90\\
   \midrule
  100 Sampling Steps& & & & \\
  Linear& 54.10& 71.59& \textbf{12.87}& \textbf{28.85}\\
  Sigmoid $\kappa=10$& \textbf{54.19}& \textbf{71.66}& 12.95& 28.86\\ %
  \rowcolor{customgreen!50} \multicolumn{5}{c}{SpaceNet 6 SAR $\rightarrow$ RGB}\\
  & mIoU $\uparrow$& mAcc $\uparrow$& FID$\downarrow$& LPIPS $\downarrow$ \\
  25 Sampling Steps& & & &\\
  Linear&64.23& 71.68& 117.30& \textbf{42.78}\\
  Sigmoid $\kappa=10$&\textbf{64.46}& \textbf{71.93}& \textbf{113.64}& 42.96\\
  \midrule
  50 Sampling Steps& & & & \\
  Linear& 63.98&71.46&119.68&\textbf{42.89}\\
  Sigmoid $\kappa=10$& \textbf{64.07}&\textbf{ 71.57}&\textbf{118.06}& 42.98\\
  \midrule
  100 Sampling Steps& & & & \\
  Linear& 63.79& 71.28& 121.28& \textbf{42.98}\\
  Sigmoid $\kappa=10$& \textbf{63.83}& \textbf{71.34}& \textbf{120.38}& 43.03\\
    \bottomrule
    \end{tabular}
}
    \caption{Sigmoid schedule vs linear schedule (preliminary results, \ours{} performances with only 100 000 training steps).}
\label{tab:scheduler_metrics}
\end{table}

\subsection{Compute time and memory footprint}

\label{supp:memory}
We report memory and times in \cref{tab:perfs}. We agree that inference time is an issue, as flow matching is slower than GANs. This is why we use a lighter distilled version of SD3's VAE (\num{0.24}s \vs \num{2.11}s for encoding-decoding). Despite relying on ODE integration, \ours transfers a batch of 256 images in 7.79s on a single A100 with 50 NFE ($\approx$\SI{30}{\milli\second}/image). 



\begin{table}[h]
    \centering
    \resizebox{\columnwidth}{!}{%
    \begin{tabular}{|l|c|c|c|} \toprule
    
         Model &  Train Mem. (GB)&  Inference Mem. (GB)&  Inference Time (s)\\ \midrule
         Pix2pix&  29.44 (64)&  14.59  (256)&  0.09 (256)\\ 
         CycleGAN&  30.75GB (12)&  14.56 (256)&  0.06 (256)\\ 
 StegoGAN& 31.67GB (8)& 24.05 (256)& 1.94 (256)\\
 UNSB& 34.00GB (12)& 0.398 (1)& 0.11 (1)\\
 \ours& 30.42GB (256)& 22.21 (256)& 7.79 (256)\\
 \bottomrule
    \end{tabular}
    }
    \footnotesize
    \caption{Memory footprints and inference times on A100 40GB. Batch sizes are indicated in brackets: measure (batch size). UNSB official implementation only supports inference batch size of 1.}
    \label{tab:perfs}
\end{table}

\begin{table*}[!t]
\centering
\setlength\tabcolsep{8pt}
\centering
\resizebox{0.99\linewidth}{!}{%
    \begin{tabular}{l cccc|cccc|cccc}
        \toprule
                 \rowcolor{customgreen!50} 
Datasets& \multicolumn{4}{c}{SpaceNet 8} & \multicolumn{4}{c}{SpaceNet 8 Germany}& \multicolumn{4}{c}{SpaceNet 8
Louisiana}\\ 
             & \multicolumn{4}{c}{Post-flood $\rightarrow$ Pre-flood} & \multicolumn{4}{c}{Post-flood $\rightarrow$ Pre-flood}& \multicolumn{4}{c}{Post-flood $\rightarrow$ Pre-flood}\\
                 & mIoU $\uparrow$& mAcc $\uparrow$& FID $\downarrow$& LPIPS $\downarrow$& mIoU $\uparrow$& Acc $\uparrow$& FID $\downarrow$& LPIPS $\downarrow$& mIoU $\uparrow$& mAcc $\uparrow$& FID $\downarrow$& LPIPS $\downarrow$ \\
            No adaptation   & 40.05& 42.40& 75.62& 63.66& 37.09& 39.08& 89.54& 63.27& 36.51& 38.85& 96.60&63.80\\
            Upper bound    & 63.10& 72.09& 00.00& 00.00& 55.27& 66.77& 00.00& 00.00& 66.91& 75.97& 00.00&00.00\\ %
            CycleGAN data-dependent& \underline{40.70}& \underline{43.35}& \textbf{54.31}& 55.70& 39.35& 41.79& \textbf{62.80}& 59.46& \underline{42.39}& \underline{45.14}& \textbf{52.80}&52.92\\ %
            CycleGAN independent& 40.64& 43.26& \textbf{52.85}& 55.17& \underline{40.34}& \underline{43.54}& 88.04& 62.01& 41.94& 44.80& 58.70&53.82\\ %
 \midrule       
            \rowcolor{customblue!50} \textbf{FlowEO}& \textbf{44.65}& \textbf{48.79}& \underline{60.32}& \textbf{45.50}& \textbf{41.27}& \textbf{45.29}& 82.74& \textbf{53.63}& \textbf{47.19}& \textbf{52.30}&  \underline{59.65}& \textbf{41.95}\\
\bottomrule
    \end{tabular}
    }
    \caption{Quantitative results on domain adaptation for weakly aligned datasets. We report both segmentation (mIoU, mAcc) and image quality metrics (FID, LPIPS) for SpaceNet 8 and its geographic subsets. CycleGAN benefits from the data-dependent coupling on SpaceNet 8 and Louisiana, despite being suited for unaligned data-translation.}
\label{tab:cycleGAN_weakly}
\end{table*}

\begin{table*}[!t]
\centering
\setlength\tabcolsep{8pt}
\centering
\resizebox{0.99\linewidth}{!}{%
    \begin{tabular}{l cccc|cccc|cccccc}
        \toprule
             \rowcolor{customgreen!50}Datasets& \multicolumn{4}{c}{Sen1Floods1} & \multicolumn{4}{c}{SpaceNet 6}& \multicolumn{6}{c}{ReBEN}\\ 
             & \multicolumn{4}{c}{SAR $\rightarrow$ Optical} & \multicolumn{4}{c}{SAR $\rightarrow$ RGB}& \multicolumn{6}{c}{SAR $\rightarrow$ Optical}\\
                 & mIoU& mAcc& FID& LPIPS & mIoU& mAcc& FID& LPIPS& $\text{AP}^\mu$& $\text{AP}^\text{M}$& $\text{F1}^\mu$&$\text{F1}^\text{M}$& FID&LPIPS  \\
   No adaptation   & 06.22& 49.72& 297.22& 84.84& 31.94& 41.01& 275.05& 79.48& 17.46& 17.43& 02.31&01.31& 339.36&85.99\\
    Upper bound    & 55.14& 71.28& 00.00& 00.00& 84.94& 90.74& 00.00& 00.00& 79.26& 65.28& 74.28&62.84& 00.00&00.00\\ %
            CycleGAN data-dependent& 42.12& 48.47& 20.97& 36.35& 50.01& 55.85& 132.75& 50.72& 26.09& 19.79& 26.93&15.75& 81.54&19.67\\
 CycleGAN independent& 44.23& 51.04& 393.88& 97.35& 51.02& 57.51& 110.90& 49.89& 24.01 & 19.88 & 28.13 & 19.77 &  78.63 & 24.08\\ %
 \midrule       
             \rowcolor{customblue!50}\textbf{FlowEO}& \textbf{54.92}&\textbf{ 69.04}& \textbf{12.96}& \textbf{29.21}& \textbf{65.07}& \textbf{72.33}& \underline{94.02}& \textbf{39.96}& \underline{37.16}& \textbf{32.14}&  \underline{36.04} & \underline{25.72}& \underline{75.80}&\textbf{15.51}\\
\bottomrule
    \end{tabular}
    }
    \caption{Quantitative results on domain adaptation for strongly aligned datasets. We report both segmentation (mIoU, mAcc) or classification (AP/F1) and image quality metrics (FID, LPIPS).  On SAR-to-optical translation datasets, CycleGAN trained with independent coupling (i.e., unaligned training) yields marginally superior performance on downstream task metrics compared to data-dependent coupling. Nonetheless, the coupling strategy does not alter its relative ranking with respect to FlowEO.}
\label{tab:CycleGAN_strongly}
\end{table*}

\section{Dataset details}
\label{sup:dataset}

For all datasets, we define three distinct splits: train, validation, and test. The training set is used to train both domain adaptation methods and predictive models. To reflect real-world scenarios -- where retraining a generative model on new data batches is impractical -- we restrict the training of image translation models to the training set. The validation set is used for hyperparameter tuning and model selection based on performance metrics, while the final reported metrics are computed on the test set.

\paragraph{SpaceNet 6} \cite{shermeyer2020spacenet} is a multimodal dataset including optical imagery (RGB bands) and SAR data (we select VV/HH/VH polarizations) at a resolution of \SI{2}{\meter\per\pixel}. From initial tiles, we crop $256\times256$ images and apply an overlap of \num{50}\% to create the training set. The segmentation masks have two different classes: background and building. We use three different splits: training ($\approx 50000$ samples), validation ($\approx 1800$), and test ($\approx 1800$) sets. 
For the optical data, we use bands $[4,3,2]$, while for the SAR data, we utilize VV, HH, and VH polarizations.

\paragraph{SpaceNet 8} \cite{9857340} is a segmentation dataset that contains pre and post-flood RGB images from Maxar for two different locations: Germany and Louisiana. The segmentation masks include three different classes: background, building, and roads. Original tiles are downsampled with a factor 2 and then cropped $256\times256$ images with an overlap of \num{70}\% to produce the training data. The final numbers of samples of each split are 5688/88/88 for Germany and 17173/244/244 for Louisiana. The full SpaceNet 8 dataset is obtained by merging the two subsets for each split.

\paragraph{Sen1Floods11} \cite{Bonafilia_2020_CVPR_Workshops} provides SAR data (Sentinel-1) and optical imagery (Sentinel-2) alongside water/non-water pixel-level annotations at a resolution of \SI{10}{\meter\per\pixel}.
Random cropping of $256\times256$ images is computed for training images, and deterministic cropping without overlap is provided for validation and test sets. It results in a total of \num{64 512} patches for training.
To match the number of SAR bands with the optical ones we duplicate the VH band, and then we use bands $[4, 3, 2]$ for optical data and VV/HH/VH polarization for SAR data.

\paragraph{BigEarthNet2 (reBEN)} \cite{clasen_reben_2025} is a multi-sensor dataset including Sentinel-1 and Sentinel-2 imagery. We used \num{237871} training patches with the multiclass annotations for both classification and domain-adaptation models training, \num{122342} for validation, and \num{119825} for testing following the original paper's splits. To match the number of SAR bands with the optical ones we duplicate the VH band, and then we use bands $[4, 3, 2]$ for optical data and VV/HH/VH polarization for SAR data. We resize the original $120\times120$ patches with bilinear interpolation to match the $256\times256$ used for the other datasets.

\section{Hyperparameters}
\label{sup:hyperparameters}
\paragraph{Pix2Pix} We train two Pix2Pix models, one translating images from $p_0$ to $p_1$ and vice versa. We use the reference PyTorch implementation available \footnote{\url{https://github.com/junyanz/pytorch-CycleGAN-and-pix2pix}} and train the models with the data-dependent coupling. We train the models with a batch size of \num{1} for \num{200000} training steps with a learning rate of \num{2e-4} and learning rate linear decay. Following the reference implementation, we use the \textit{LSGAN} \cite{mao2017least} adversarial loss. We deviate from the default hyperparameters for $\lambda_\text{L1}$, which we decrease from \num{100} to \num{10} to fix blurry image generation issues on ours datasets. The generator is a 9-blocks ResNet and we use the PatchGAN discriminator \cite{isola2017image} with instance normalization.

\paragraph{CycleGAN} The implementation of CycleGAN follows the same hyperparameters set as the Pix2Pix mentioned above. We train the models with a batch size of \num{1} for \num{200000} training steps with a learning rate of \num{2e-4} and learning rate linear decay. We keep $\lambda_\text{L1} = 100$ since it does not negatively impact the training or the generated images' quality. We used the same network architectures as for Pix2Pix.

\paragraph{StegoGAN} While the StegoGAN models use two generators, translating respectively from domain $\mathcal{X}_0$ to $\mathcal{X}_1$ and vice versa, the training process is asymmetrical. Thus, we trained two different models for each dataset, using the official implementation\footnote{\url{https://github.com/sian-wusidi/StegoGAN}}.
We use \textit{LSGAN} adversarial loss, instance normalization, and train the model for \num{200000} iterations with a learning rate of \num{2e-4}. We select the set of loss weightings used for the GoogleMismatch dataset in the original paper: $\lambda_A=10$, $\lambda_B=10$,$\lambda_A=10$, $\lambda_\text{id}=0.5$, $\lambda_\text{cycle}=0.5$ and $\lambda_\text{reg}=0.3$ for the mask regularization loss. Note that this last value is similar for all remote sensing datasets used in StegoGAN: $\lambda_\text{cycle}=0.5$ for GoogleMismatch and $\lambda_\text{cycle}=0.3$ for PlanIGN.  The generator is a 9-blocks-Resnet and we use the PatchGAN discriminator \cite{isola2017image} with instance normalization.

\paragraph{UNSB} Schrödinger bridges map two arbitrary distributions with forward and backward stochastic processes. Nevertheless, UNSB leverages an adversarial loss on $p_1$ making the training asymmetrical. Thus we train two different models, translating respectively from domain $\mathcal{X}_0$ to $\mathcal{X}_1$ and vice versa. We use the official implementation \footnote{\url{https://github.com/cyclomon/UNSB}} and train the models for \num{200000} iterations with a learning rate of \num{2e-4}. We use the proposed set of hyperparameters: $\lambda_\text{GAN} = 1$, $\lambda_\text{NCE} =1$, $\lambda_\text{SB} =1$. We use the same architectures as the other methods, namely 9-blocks-Resnet and PatchGAN discriminator with instance normalization. We use \num{5} sampling steps at inference, following original paper guidelines.

\paragraph{Diffusion Bridges} Diffusion bridges establish mappings between arbitrary distributions via forward and backward stochastic processes. We adopt the formulation of \cite{bortoli2024schrodinger} and train the models for \num{200000} iterations using an $x_1$-prediction objective, with a batch size of 32 and a learning rate of \num{2e-4}. The UNet backbone follows the same design as \ours, but is adapted to operate directly on image inputs rather than latent representations. Inference is performed with \num{50} sampling steps, consistent with \ours.

\section{CycleGAN with unaligned training}

CycleGAN is a data-to-data translation framework originally designed to handle unaligned datasets through its cyclical loss. However, in the context of pre- and post-disaster datasets, we observe that CycleGAN benefits from the availability of co-registered pairs (data-dependent coupling improves segmentation metrics) (\Cref{tab:cycleGAN_weakly}). For SAR-to-optical translation, the use of unpaired datasets can offer certain advantages, though the performance gains are marginal and do not alter its relative ranking compared to our method (\Cref{tab:CycleGAN_strongly}).

\section{Additional quantitative results on reBEN}

We include in \Cref{tab:perfs_pix2pix_reben} a detailed comparison of Pix2pix and FlowEO on the ReBEN SAR-to-Optical domain adaptation dataset. It reveals that Pix2pix exhibits a pronounced bias toward forest classes (\textit{Coniferous forest} and \textit{Mixed forest} classes), which are disproportionately represented relative to other categories. This class imbalance inflates micro-averaged metrics, thereby explaining the discrepancy in ranking between FlowEO and Pix2pix under micro- versus macro-averaging.

\begin{table*}[h]
    \centering
    \resizebox{\linewidth}{!}{%
    \begin{tabular}{lccccrr}
                                                                                       & \multicolumn{1}{l}{\cellcolor[HTML]{D9EAD3}Pix2Pix} & \multicolumn{1}{l}{\cellcolor[HTML]{D9EAD3}FlowEO} & \multicolumn{1}{l}{\cellcolor[HTML]{D9EAD3}Pix2Pix} & \multicolumn{1}{l}{\cellcolor[HTML]{D9EAD3}FlowEO} & \multicolumn{1}{l}{\cellcolor[HTML]{D9EAD3} \#test samples} & \multicolumn{1}{l}{\cellcolor[HTML]{D9EAD3}Proportions} \\
                                                                                       & \multicolumn{2}{c}{AP}& \multicolumn{2}{c}{F1}& \multicolumn{1}{l}{}               & \multicolumn{1}{l}{}            \\
Macro metric $M$                                                                                  & 27.88                                               & 32.14                                              & 25.79                                               & 25.72                                              & \multicolumn{1}{l}{}               & \multicolumn{1}{l}{}            \\
Micro metric $\mu$                                                                                  & 41.09                                               & 37.16                                              & 43.93                                               & 36.04                                              & \multicolumn{1}{l}{}               & \multicolumn{1}{l}{}            \\
                                                                                       & \multicolumn{1}{l}{}                                & \multicolumn{1}{l}{}                               & \multicolumn{1}{l}{}                                & \multicolumn{1}{l}{}                               & \multicolumn{1}{l}{}               & \multicolumn{1}{l}{}            \\
Industrial or commercial units                                                         & 13.79                                               & 25.43                            & 22.47                                               & 34.09                                              & 2018                               & 0.0058                          \\
Arable land                                                                            & 64.25                                               & 73.77                                              & 62.05                                               & 69.89                                              & 50052                              & 0.1446                          \\
Permanent crops                                                                        & 6.69                                                & 11.42                                              & 05.02                                                & 12.19                                              & 5710                               & 0.0165                          \\
Pastures                                                                               & 35.01                                               & 42.38                                              & 22.84                                               & 36.22                                              & 26722                              & 0.0772                          \\
Complex cultivation patterns                                                           & 24.70                                               & 30.58                                              & 08.06                                                & 36.28                                              & 22078                              & 0.0638                          \\
Land principally occupied by agriculture, with significant areas of natural vegetation & 31.46                                               & 35.99                                              & 33.35                                               & 30.75                                              & 29846                              & 0.0862                          \\
Agro-forestry areas                                                                    & 22.62                                               & 44.25                                              & 05.55                                                & 18.56                                              & 9942                               & 0.0287                          \\
Broad-leaved forest                                                                    & 32.76                                               & 41.63                                              & 22.76                                               & 20.68                                              & 36377                              & 0.1051                          \\
Coniferous forest                                                                      & 54.65                                               & 54.95                                              & 57.82                                               & 30.66                                              & 39043                              & 0.1128                          \\
Mixed forest                                                                           & 52.64                                               & 49.57                                              & 58.93                                               & 29.07                                              & 44284                              & 0.1280                          \\
Natural grassland and sparsely vegetated areas                                         & 01.57                                                & 02.30                                               & 00.08                                                & 02.32                                               & 2211                               & 0.0064                          \\
Moors, heathland and sclerophyllous vegetation                                         & 03.74                                                & 05.31                                               & 02.39                                                & 02.70                                               & 3759                               & 0.0109                          \\
Transitional woodland, shrub                                                           & 43.34                                               & 44.00                                              & 45.68                                               & 29.54                                              & 40523                              & 0.1171                          \\
Beaches, dunes, sands                                                                  & 00.92                                                & 00.75                                               & 03.88                                                & 02.29                                               & 152                                & 0.0004                          \\
Inland wetlands                                                                        & 05.26                                                & 04.98                                               & 08.96                                                & 09.24                                               & 4519                               & 0.0131                          \\
Coastal wetlands                                                                       & 00.09                                                & 00.09                                               & 00.28                                                & 00.11                                               & 117                                & 0.0003                          \\
Inland waters                                                                          & 33.79                                               & 34.17                                              & 26.53                                               & 25.78                                              & 16846                              & 0.0487                          \\
Marine waters                                                                          & 69.16                                               & 68.78                                              & 66.72                                               & 55.48                                              & 11854                              & 0.0343                         
\end{tabular}
}
    \footnotesize
    \caption{Performance comparison of Pix2pix and FlowEO on the ReBEN SAR-to-Optical domain adaptation dataset. Pix2pix shows a strong bias toward forest classes, which are overrepresented relative to other categories. The high performance on these dominant classes inflates micro-averaged metrics, accounting for the difference in ranking between FlowEO and Pix2pix under micro- versus macro-averaging.}
    \label{tab:perfs_pix2pix_reben}
\end{table*}

\section{Additional qualitative results}

\subsection{Qualitative classification results on reBEN}

We provide here qualitative domain adaptation results for reBEN, with transferred images for baselines and \ours{} and predicted labels shown in \Cref{fig:reben_qualitative}. As for the segmentation tasks, this underlines both the visual quality of the generated images by \ours and the accuracy of the predictions by the pre-trained classification model on the adapted images. In addition to the generated optical images, we show the top-3 predicted classes, \ie the 3 classes with the highest probabilities predicted by the classification model $C^*_1$.

\begin{figure*}[ht]
\centering
\input{figures/visual_reben}
\caption{Qualitative comparison of domain adaptation methods on the reBEN dataset, for multiclass classification. The first column represents the source domain image $x_0$, the second depicts the weakly or strongly aligned $x_1$, and the others display the images generated by the different methods. Below each image generated, we provide the corresponding top-3 predicted classes by the classification model $C_1$. For the reference image, we display all the class labels. \ours{} outperforms other methods in both class preservation and image quality.}
\label{fig:reben_qualitative}
\end{figure*}

\subsection{Additional image generation results}

\begin{figure*}[ht]
\centering
\input{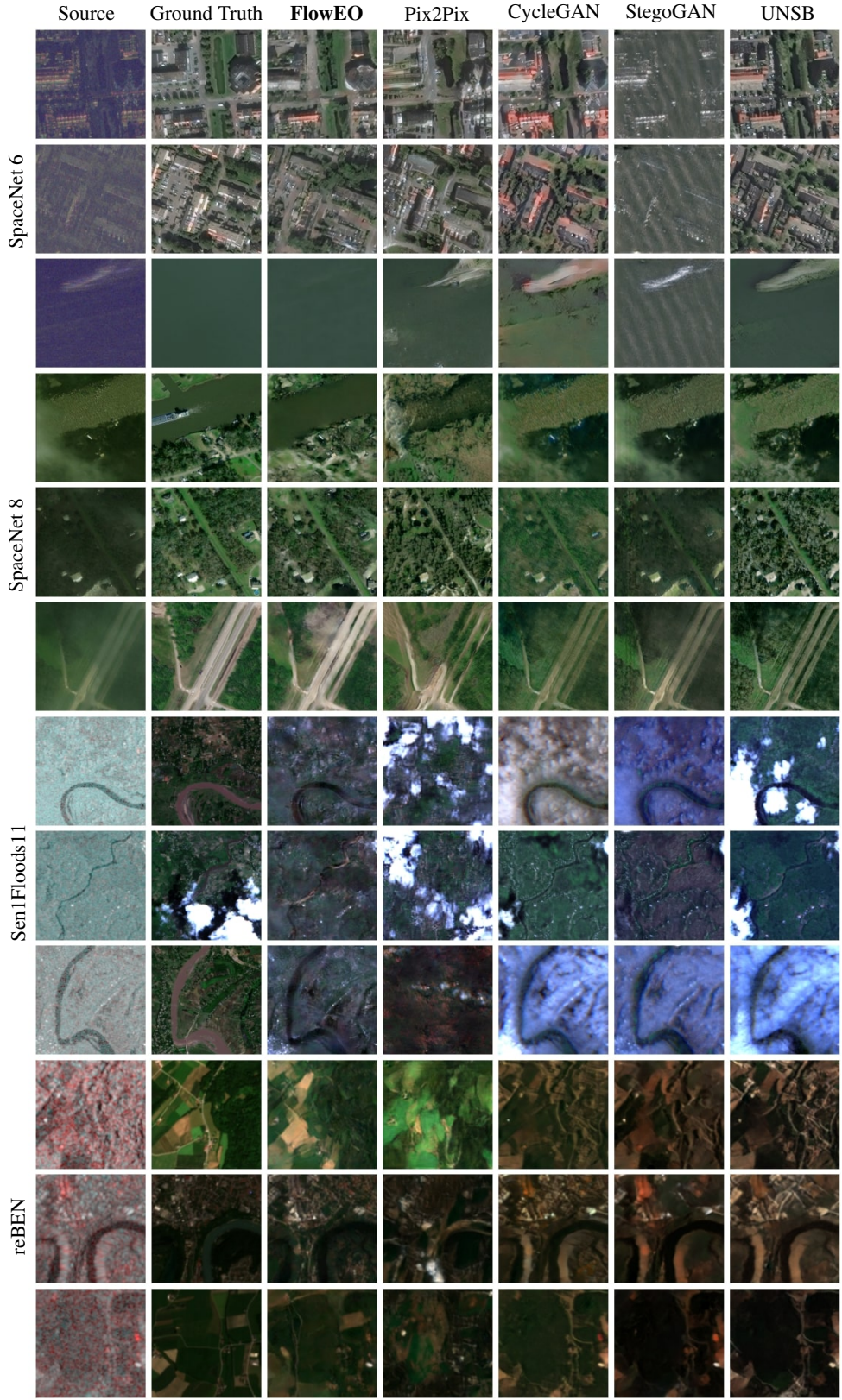}
\caption{\ours{} generates the highest-quality images while maintaining semantic consistency during the transfer process. In the third row, we observe that our method demonstrates greater robustness to the geometric artifacts present in SAR imagery. Additionally, we note that it successfully learns to map flood-disturbed water states to a more natural appearance (fourth row).}
\label{fig:additional_qualitative}
\end{figure*}

We provide in \Cref{fig:additional_qualitative} additional image generation results for a more exhaustive assessment of our image translation approach.
We can observe that \ours tends to better capture the color range of the reference images, avoid hallucinations, and better reconstruct the scene geometry. In particular, note that \ours is robust to changes between the source and target images, \eg clouds and boats that have moved. Interestingly, this shows the potential of flow matching for inverse problems in Earth observation, such as cloud removal.

\end{document}